\documentclass[runningheads]{lncs}

\usepackage{graphicx}
\usepackage{tikz,pgfplots}

\usepackage{amsmath}

\usepackage{cite}

\usepackage{xcolor}
\usepackage{latexsym}
\usepackage{hyperref}
\usepackage{float}
\usepackage{amsthm}
\usepackage{amssymb}
\usepackage{pifont}
\newcommand{\cmark}{\ding{51}}
\newcommand{\xmark}{\ding{55}}

\usepackage{pgfplots}
\usepgfplotslibrary{colorbrewer}
\pgfplotsset{compat = 1.15, cycle list/Set1-8} 
\usetikzlibrary{pgfplots.statistics, pgfplots.colorbrewer} 
\usepackage{pgfplotstable}
\usepackage{filecontents}

\newcommand{\MI}{\textsf{MI}}

\newcommand{\Free}{\textsf{Free}}

\newcommand{\rb}{\mathcal{B}}
\newcommand{\ms}{\mathcal{M}}
\newcommand{\F}{\mathcal{F}}
\newcommand{\R}{\mathcal{R}}
\newcommand{\atoms}{\ensuremath{\mathcal{A}}}
\newcommand{\lang}{\ensuremath{\mathcal{L}}}

\newcommand{\allrbs}{\ensuremath{\mathbb{B}}}
\newcommand{\allrs}{\ensuremath{\mathbb{R}}}
\newcommand{\allmss}{\ensuremath{\mathbb{M}}}

\newcommand{\culp}{\ensuremath{\mathcal{C}}}
\newcommand{\inc}{\ensuremath{\mathcal{I}}}

\newcommand{\posRealInf}{\ensuremath{\mathbb{R}^{\infty}_{\geq 0}}}

\newcommand{\incmi}{\ensuremath{\mathcal{I}_{\mathsf{MI}}}}

\begin{document}

\title{Measuring Inconsistency over Sequences of Business Rule Cases\thanks{This  research  is  part  of  the  research  project  ”Handling  Inconsistencies  in Business Process Modeling“, which is funded by the German Research Association (reference number:  DE1983/9-1).}}

\author{Carl Corea\inst{1} \and
Matthias Thimm\inst{2} \and
Patrick Delfmann\inst{1}}
\authorrunning{C. Corea et al.}

\institute{Institure for Information Systems Research, University of Koblenz-Landau \and
Institute for Web-Science and Technologies, University of Koblenz-Landau
\email{\{ccorea,thimm,delfmann\}@uni-koblenz.de}}
\maketitle              
%

\begin{abstract}
In this report, we investigate (element-based) inconsistency measures for multisets of business rule bases. Currently, related works allow to assess individual rule bases, however, as companies might encounter thousands of such instances daily, studying not only individual rule bases separately, but rather also their interrelations becomes necessary, especially in regard to determining suitable re-modelling strategies. We therefore present an approach to induce multiset-measures from arbitrary (traditional) inconsistency measures, propose new rationality postulates for a multiset use-case, and investigate the complexity of various aspects regarding  multi-rule base inconsistency measurement.

\end{abstract}

%
%
\section{Introduction}\label{sec:introduction}
\begin{figure}[H]
    \centering
    \includegraphics[width=.73\columnwidth]{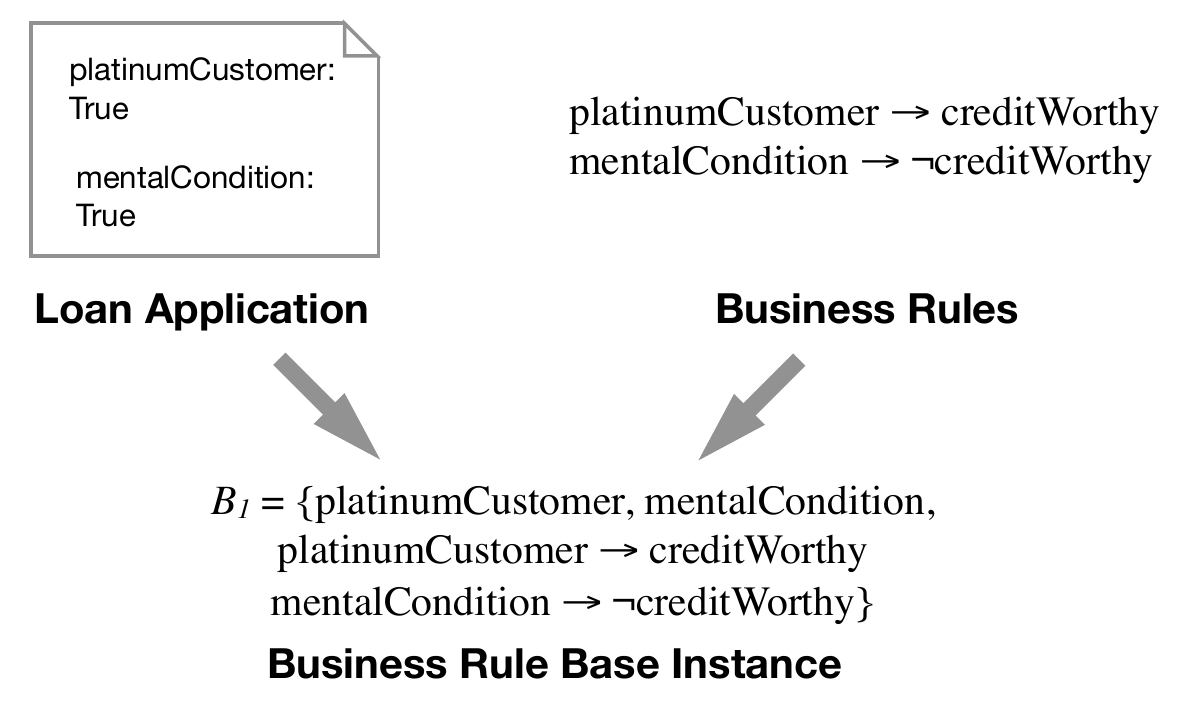}
    \caption{Exemplary business rule base instance $\rb_1$.}
    \label{fig:exemplaryRuleBase}
\end{figure}

In the context of Business Process Management, \emph{business rules} are used as a central artifact to govern the execution of company activities \cite{graham:2007business}. To this aim, business rules are modelled to capture (legal) regulations as a declarative business logic. Then, given a new process instance (denoted as a \emph{case}), instance-dependent facts are evaluated against the set of business rules for reasoning at run-time. For example, consider the following set of business rules in Figure \ref{fig:exemplaryRuleBase} (we will formalize syntax and semantics later) with the intuitive meaning that we have two rules stating that 1) platinum customers are credit worthy, and 2) customers with a mental condition are not credit worthy. Then, given a new customer case, in the example a new loan application, the facts set is evaluated against the rule set and the resulting rule base $\rb_1$ can be used to reason about the customer case.

The observant reader might have noticed, that the shown example yields an inconsistency, i.e., the contradictory conclusions $\mathit{creditWorthy, \neg creditWorthy}$. In fact, this is a current problem for companies, which can result from modelling errors in the business rules, or unexpected (case-dependent) facts. This problem has widely been acknowledged and has been addressed by a series of recent works, cf. e.g. \cite{corea:2019b,diciccio:2017resolving,corea2020towards}. 

While existing results allow to handle inconsistencies in a \emph{single} business rule base instance as shown above, in practice, companies often face thousands of such instances daily. For example, the retailer Zalando reported that 37 million cases were executed in the first quarter of 2020 alone\footnote{\url{https://zln.do/2SFRnfC}}. As we will show in this work, considering not only single rule bases individually, but rather the entirety of all cases and their \emph{interrelations}, can yield valuable insights, especially in regard to inconsistency resolution. For example, consider the following rule set, and assume there were four customer cases (with respective case-dependent facts), yielding the set of business rule cases $\ms_1$ shown in Figure~\ref{fig:exemplaryInstances}:

\begin{figure}[H]
    \centering
    \includegraphics[width=.83\columnwidth]{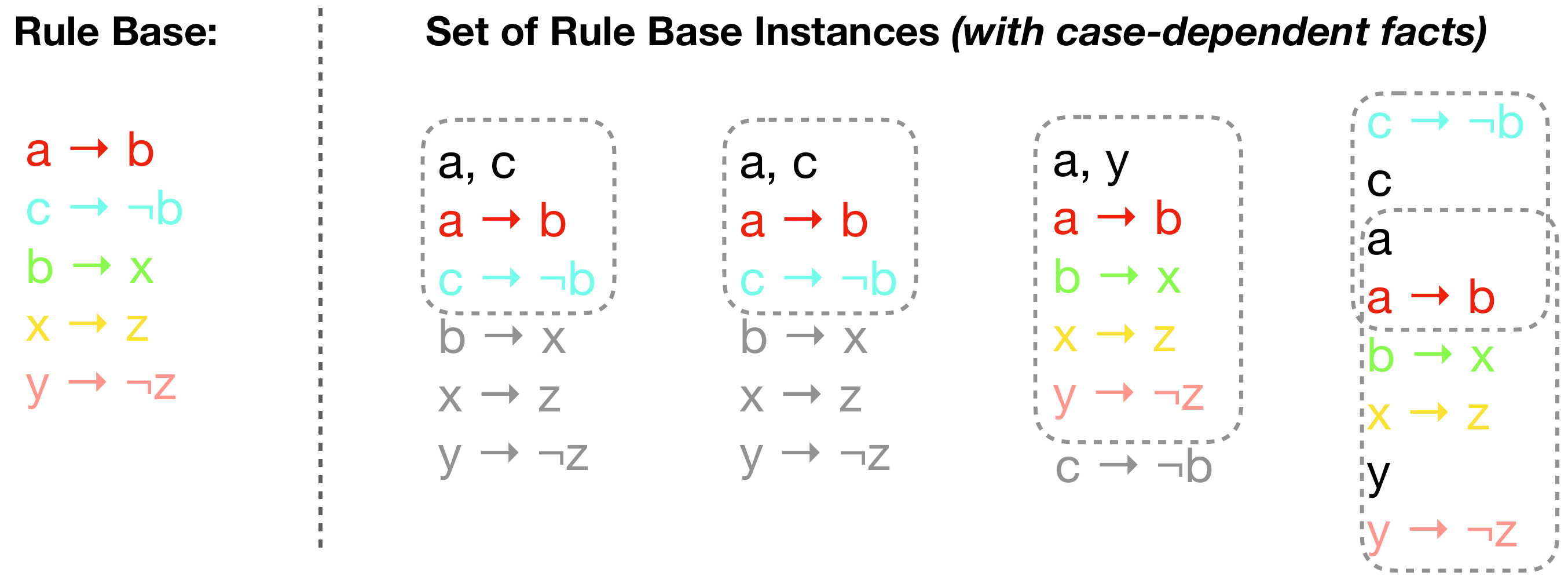}
    \caption{Exemplary rule base instances, constructed over a seq. of case-dependent facts.}
    \label{fig:exemplaryInstances}
\end{figure}

When auditing such an overview of rule base instances, two main questions are of interest from a business rules management perspective:
\begin{enumerate}
    \item How inconsistent in general was the entirety of process executions?
    \item Which specific rules were responsible for these inconsistencies from a global perspective?
\end{enumerate}
Towards question~1, recent results studying inconsistency in sets of knowledge bases can easily be adapted to quantify the overall degree of inconsistency (cf. Section~2). While this is a beneficial step for companies, question~2 can however be seen as of much higher importance in the scope of improving business rules. Pin-pointing the culprits of inconsistency is an essential challenge for determining suitable resolution and re-modelling strategies. Here, new methods are needed that support companies in assessing which individual rules are highly problematic from a global perspective. 
For instance, in Figure \ref{fig:exemplaryInstances}, the rule $a\rightarrow b$ is part of all inconsistencies and can therefore be seen as highly problematic. In this work, we therefore introduce novel means for an element-based assessment of inconsistency over a set of rule base instances by extending results from the field of \emph{inconsistency measurement} \cite{Thimm:2019d}. Here, our contribution is as follows:

We present a novel approach for inducing element-based quantitative measures for multisets of business rule bases, allowing to pin-point problematic business rules from a global perspective (Section 3). Here, we also propose postulates that should be satisfied by respective measures for this use-case and analyze the proposed means w.r.t. these postulates. We implement our approach and perform run-time experiments with real-life data-sets, and also examine the complexity of central aspects regarding inconsistency measurement in multisets of business rule bases (Section 4). We present preliminaries in Section 2 and conclude in Section 5. Proofs for technical results are provided in a supplementary document\footnote{\url{https://bit.ly/2V2sIDw}}.

%
%
\section{Preliminaries}\label{sec:preliminaries}

\textbf{Business Rule Bases.} In this work, we consider a basic (monotonic) logic programming language to formalise business rule bases. A \emph{(business) rule base} is then constructed over a finite set $\atoms$ of atoms, with $\lang$ being the corresponding set of literals, with a rule base $\rb$ being a set of rules $r$ of the form
\begin{align}\label{eq:rule}
r\,:\quad  l_{1},\ldots,l_{m} \rightarrow l_{0}.
\end{align}
with every $l_i\in\lang$. Let $\allrbs$ denote  all such rule bases. Also, we denote $head(r)=l_{0}$ and $body(r)=\{l_{1},\ldots,l_{m} \}$. If $body(r)=\emptyset$,  $r$ is called a \emph{fact}. 
For a rule base $\rb$, we denote $\F(\rb)\subseteq\rb$ as the facts in $\rb$ and $\R(\rb)\subseteq\rb$ as the rules in $\rb$. 

\begin{example}
We recall the business rule base $\rb_1$. Then we have
\begin{align*}
	\F(\rb_1) & = \{mentalCondition, platinumCustomer\}\\
	\R(\rb_1) & = \{platinumCustomer \rightarrow creditWorthy,\\
	& \qquad mentalCondition \rightarrow \neg creditWorthy\}.
\end{align*}
\end{example}

A set of literals $M$ is called \emph{closed} w.r.t. $\rb$ if it holds that for every rule of the form \ref{eq:rule}: if $l_{1},\ldots,l_{m}\in M$ then $l_{0}\in M$. The \emph{minimal model} of a rule base $\rb$ is the smallest closed set of literals (w.r.t. set inclusion). A set $M$ of literals is called consistent if it does not contain both $a$ and $\neg a$ for an atom $a$. We say a rule base $\rb$ is consistent if its minimal model is consistent. If $\rb$ is not consistent, we say $\rb$ is inconsistent, denoted as $\rb\models\perp$.

As discussed in the introduction, $\rb_1$ is not consistent. To assess inconsistency, the field of \emph{inconsistency measurement} \cite{Grant:2018} has evolved, which studies quantitative measures to assess the severity of inconsistency. An inconsistency measure \cite{Grant:2018,Thimm:2019d} is a function $\inc: \allrbs \rightarrow \posRealInf$, where the semantics of the value are defined such that a higher value reflects a higher degree, or severity, of inconsistency. A basic inconsistency measure is the $\incmi$ inconsistency measure, which counts the number of minimal inconsistent subsets $\MI$ of a rule base $\rb$, defined via
\begin{align*}
	\MI(\rb) = \{M \subseteq \rb \mid M \models \perp, \forall M' \subset M : M' \not\models \perp \}. 
\end{align*}
\begin{example}We recall $\rb_1$. Then we have
\begin{align*}
    \MI(\rb_1) &= \{M_1\}\\
	M_{1}	&= \{platinumCustomer,\\
	        & \qquad platinumCustomer \rightarrow creditWorthy,\\
			& \qquad mentalCondition,\\
			& \qquad mentalCondition \rightarrow \neg creditWorthy\},
\end{align*}
consequently, $\incmi(\rb_1)=1$.
\end{example}

As the concept of a ''severity`` of inconsistency is not easily characterisable, numerous inconsistency measures have been proposed, see \cite{Thimm:2019d} for an overview. To guide the development of inconsistency measures, various rationality postulates have been proposed, cf. \cite{thimm:2017compliance} for an overview. For example, a widely agreed upon property is that of \emph{consistency}, which states that an inconsistency measure should return a value of 0 w.r.t. a rule base $\rb$ iff $\rb$ is consistent.
As mentioned, various other postulates exist and we will revisit some of them later when introducing culpability measures for multisets of rule bases.

\textbf{Measuring Overall Inconsistency in Multisets of Business Rule Bases.}
In this work, we are not only interested in measuring inconsistency in single business rule bases, but rather in a series of corresponding business rule base instances. As motivated in the introduction, companies currently apply a set of business rules in order to assess a stream of (case-dependent) fact sets. Therefore, given a stream of fact sets $f = \F_1,...,\F_n$, we consider multisets of business rule bases which are constructed by matching the individual fact sets in $f$ to a shared rule set $\R$. To clarify, a multiset of rule bases is an n-tuple $\ms=(\{\F_1\cup\R\},...,\{\F_n\cup\R\})=(\rb_1, ..., \rb_n)$.  Let $\allmss$ denote all such multisets.

An initial question for companies is to gain an overview of the overall inconsistency w.r.t. all business rule base instances in $\ms$. For that, we define an inconsistency measure for a multiset of rule bases as follows.
\begin{definition}[Multi-$\rb$ Inconsistency Measure]
An inconsistency measure for a multiset of rule bases is a function $m: \allmss \rightarrow \posRealInf$.
\end{definition}
In other words, an inconsistency measure for a multiset of rule bases is a function that assigns a non-negative numerical value to an n-tuple of rule bases. Similar to classical inconsistency measures, the intuition is that a higher value reflects a higher degree of inconsistency of the multiset of rule bases. For simplicity, we refer to such measures as multi-rb measures where appropriate.

For the intended use-case of gaining insights about the severity of inconsistency regarding the entirety of process instances, existing inconsistency measures can be adapted to induce multi-rb measures via a summation.
\begin{definition}[$\Sigma$-induced multi-rb Measure]
Given an inconsistency measure $\inc$ and a multiset of rule-bases $\ms$, the $\Sigma$-induced multi-rb measure $m_\inc^\Sigma$ is defined as $m_\inc^\Sigma: \allmss \rightarrow \posRealInf$ with $m_\inc^\Sigma(\ms)=\sum_{B \in \ms} \inc(B)$.
\end{definition}


\begin{example}
We recall the introduced $\MI$-inconsistency measure $\incmi$. Correspondingly, given a multiset of rule bases $\ms$, $\incmi$ can be used to $\Sigma$-induce the multi-rb measure $m_{\incmi}^{\Sigma}(\ms) = \sum_{B \in \ms} \incmi(B)$. Considering again the exemplary multiset $\ms_1$ of business rules from Figure \ref{fig:exemplaryRuleBase}, with cases $b_1-b_4$, we thus have $m_{\incmi}^{\Sigma}(\ms_1) = \incmi(b_1) + ... + \incmi(b_4) = 1+1+1+2 = 5$.
\end{example}
Note that the approach in \cite{potyka2018measuring}, who---roughly speaking---measures inconsistency in a multiset of knowledge bases by performing a multiset union on all sets and then measuring inconsistency on this union, is not applicable for our use-case, as we are not interested in the disagreement between the individual instances, but rather want to gain an overview of inconsistencies in all instances.


While the above discussion showed how existing means can be used to measure the \emph{overall} degree of inconsistency for a multiset of rule bases, in the following, we develop techniques for an \emph{element-based} assessment of inconsistency over a multiset of rule bases.

%
%
\section{Culpability Measures for Multisets of Business Rule Bases}\label{sec:culpabilityMeasuresMain}
In the field of inconsistency measurement, a culpability measure $\culp$ \cite{hunter2010measure} is a function that assigns a non-negative numerical value to elements of a rule base. This quantitative assessment is also referred to as an inconsistency value. Again, the intuition is that a higher inconsistency value reflects a higher blame, that the specific element carries in the context of the overall inconsistency. In this section, we investigate culpability measures that can assess the blame that a rule carries in the context of the overall multiset inconsistency.

\subsection{Baseline measures and basic properties}\label{sec:culpabilityMeasuresBaseline}
Given a multiset of business rule bases $\ms$, let $\R(\ms)$ denote the shared rule set of the respective business rule bases in $\ms$. Furthermore, let $\allrs_\allmss$ denote the set of all possible rules that can appear in these shared rule sets. Then, a culpability measure for a multiset of rule bases is defined as follows.
\begin{definition}[Multi-$\rb$ Culpability Measure]
A culpability measure for a multiset of rule bases is a function $\culp^m:\allmss \times \allrs_\allmss \rightarrow \posRealInf$.
\end{definition}

Similar to $\Sigma$-induced inconsistency measures, existing culpability measures can be exploited to entail $\Sigma$-induced culpability measures.

\begin{definition}[$\Sigma$-induced multi-rb culpability measure]
Given a culpability measure $\culp$, a multiset of rule-bases $\ms$ and a rule $r\in\R(\ms)$, a $\Sigma$-induced multi-rb culpability measure $m_{\culp}^{\Sigma}$ is defined as $m_{\culp}^{\Sigma}:\allmss \times \allrs_\allmss \rightarrow \posRealInf$ with $m_{\culp}^{\Sigma}(\ms,r)=\sum_ {B\in\ms}\culp(B,r)$.
\end{definition}

Two baseline measures proposed in \cite{hunter2008measuring} are the $\culp_D$ and $\culp_{\#}$ measures.
\begin{definition}
Let a rule base $\rb$ and a rule $r\in\rb$, then
\begin{itemize}
    \item $\culp_D(\rb,r) = 
        \begin{cases}
            1 & \text{if } \exists M \in \MI(\rb):r\in M \\
            0 & \, \text{otherwise}
        \end{cases}$
    \item $\culp_{\#}(\rb,r) = |\{M \in \MI(\rb) \mid r \in M\}|$
\end{itemize}
\end{definition}

Using $\Sigma$-induction, we can use these baseline culpability measures to entail the multi-rb culpability measures $m_{\culp_D}^{\Sigma}$ and $m_{\culp_{\#}}^{\Sigma}$. 

\begin{example}
We recall the multiset of rule bases $\ms_1$ from Figure \ref{fig:exemplaryInstances}. For the shown rule $a\rightarrow b$, we have that
\begin{align*}
    m_{\culp_D}^{\Sigma}(\ms_1, a\rightarrow b) & = 4\\
    m_{\culp_{\#}}^{\Sigma}(\ms_1, a\rightarrow b) & = 5
\end{align*}
\end{example}

Regarding multi-rb culpability measures, we propose the following rationality postulates based on an application of  postulates for traditional culpability measures \cite{hunter2010measure}. For that, we consider a multiset of business rules $\ms$ and a rule $r\in\R(\ms)$. Also, we define a rule $r\in\R(\ms)$ as a \emph{free formula} if $r \notin M, \forall M \in \bigcup_{b\in\ms}\MI(b)$. We denote the set of all free formulas of $\R(\ms)$ as $\Free(\ms)$. We then propose the following postulates.
\begin{description}
	\item[\emph{Rule Symmetry} (\textsf{RS})] $\culp^m(\ms,r)=\culp^m((\rb_{1},...\rb_n),r)$, for any permutation of the order of $\rb_1$ to $\rb_n$.
	\item[\emph{Rule Minimality} (\textsf{RM})] if $r\in \Free(\ms)$, then $\culp^m(\ms,r)=0$.
\end{description}
The first postulate states that the order of rule bases in the multiset should not affect the inconsistency value of an individual rule. The second postulate states that the inconsistency value of a rule is zero if this rule is a free formula w.r.t. the multiset of business rule bases.

\begin{proposition}
$m_{\culp_D}^{\Sigma}$ and $m_{\culp_{\#}}^{\Sigma}$ satisfy \textsf{RS} and \textsf{RM}.
\end{proposition}

The second postulate was adapted from a postulate for traditional culpability measures, namely 
\begin{description}
	\item[\emph{Minimality} (\textsf{MIN})]Let a rule $r\in\rb$, if $r\not\in M, \forall M \in \MI(\rb)$, then the inconsistency value of r is zero.
\end{description}
As this is a commonly satisfied postulate, this allows for a generalization of the previous proposition.
\begin{proposition}
Any $\Sigma$-induced multi-rb culpability measure satisfies \textsf{RS}. Given a culpability measure $\culp$ satisfying \textsf{MIN}, any multi-rb culpability measure $\Sigma$-induced via $\culp$ satisfies \textsf{RM}.
\end{proposition}

In the following, given a multiset of business rules $\ms$ and a multi-rb culpability measure $\culp^m$, we consider all rules of $\R(\ms)$ as a vector $(r_1,...r_n)$, and denote $V^{\culp^m}(\ms)$ as the vector of corresponding multi-rb culpability values of all rules in $\R(\ms)$ w.r.t. $\culp^m$, i.e., $V^{\culp^m}(\ms) = (\culp^m(\ms,r_1),...,\culp^m(\ms,r_n))$. Next, let $\hat{V}^{\culp^m}(\ms)=\mathit{max}_{r\in\R(\ms)}(\culp^m(\ms,r))$ denote the largest multi-rb culpability value w.r.t. $\culp^m$ for all rules. Last, we denote adding a rule $r$ to the shared rule set $\R(\ms)$ of a multiset $\ms$ as $\ms\cup \{r\}$ by a slight missuse of notation, i.e., given $\ms = (\rb_1,...,\rb_n), \ms\cup \{r\}= (\rb_1 \cup \{r\},...,\rb_n\cup \{r\})$. 
This allows to adapt some further desirable properties.
\begin{description}
	\item[\emph{Multiset Consistency} (\textsf{CO})] $\hat{V}^{\culp^m}(\ms)=0$ iff $\nexists \rb\in\ms:\rb\models\perp$.
	\item[\emph{Multiset Monotony} (\textsf{MO})] Let a multiset of business rule bases $\ms$ and a rule r, $\hat{V}^{\culp^m}(\ms \cup \{r\}) \geq \hat{V}^{\culp^m}(\ms)$
	\item[\emph{Multiset Free formula independence} (\textsf{IN})] If a rule r is a free formula of $(\ms\cup\{r\})$, then $\hat{V}^{\culp^m}(\ms \cup r) = \hat{V}^{\culp^m}(\ms)$
\end{description}
The first property states that the largest multi-rb culpability value for a rule can only be zero if all business rule bases of the multiset are consistent. The second property demands that adding a rule to the shared rule set can only increase the culpability values. Similar to this property, the third postulate demands that adding a free formula to the shared rule set does not alter the culpability values.

\begin{proposition}
$m_{\culp_D}^{\Sigma}$ and $m_{\culp_{\#}}^{\Sigma}$ satisfy \textsf{CO}, \textsf{MO} and \textsf{IN}.
\end{proposition}

Next to the introduced baseline culpability measures $\culp_D$ and $\culp_{\#}$, various other culpability measures have been proposed (cf. \cite{mcareavey2014computational}), which could also be used to $\Sigma$-induce multi-rb culpability measures. While an analysis of such measures w.r.t. the introduced postulates could be interesting, we refrain from such a specific analysis and rather show a more generalized approach in the following section, namely how arbitrary inconsistency measures can be used to induce multi-rb culpability measures using Shapley inconsistency values.

%
%
\subsection{(Adjusted) Shapley Inconsistency Values for Multi-RB measures }\label{sec:culpabilityMeasuresShapley}

Next to designing specific culpability measures, an important approach in element-based analysis is to decompose the assessment of inconsistency measures (in order to derive corresponding culpability measures) by means of Shapley inconsistency values \cite{hunter2010measure}. Given an inconsistency measure $\inc$ and a rule base $\rb$, the intuition is that the overall blame mass $\inc(\rb)$ is distributed amongst all elements in $\rb$, by applying results from game theory. The advantage of this approach is that arbitrary inconsistency measures can be applied to derive a corresponding element-based assessment. The amount of blame that an individual element is assigned relative to $\inc(\rb)$ is also referred to as the \emph{payoff}. 

\begin{definition}[Shapley Inconsistency value \cite{hunter2010measure}]
    Let $\inc$ be an inconsistency measure, $\rb$ be a rule base and $\alpha \in \rb$. Then, the Shapley inconsistency value of $\alpha$ w.r.t. $\inc$, denoted $S_\alpha^{\inc}$ is defined via
    \begin{align*}
        S_\alpha^{\inc}(\rb) = \sum_{B\subseteq\rb}\frac{(b-1)!(n-b)!}{n!}(\inc(\rb)-\inc(\rb\setminus \alpha))
    \end{align*}
    where $b$ is the cardinality of $B$, and $n$ is the cardinality of $\rb$.
\end{definition}
\begin{example}\label{ex:shapleyIgnoringFactContradictions}
Consider the rule base $\rb_2 = \{a,a\rightarrow b,a\rightarrow \neg b\}$. Then, for the Shapley inconsistency values w.r.t. $\incmi$, for all elements $e$ in $\rb$ we have that $S^{\incmi}_{e}(\rb_2) = \frac{1}{12}+\frac{1}{4} = \frac{1}{3}$. 
\end{example}
As shown in Example \ref{ex:shapleyIgnoringFactContradictions}, all elements were assigned an equal payoff. This makes sense w.r.t. $\incmi$ if all elements in the rule base are considered with equal importance. However, in our setting, knowledge contained in rule bases is distinguished into facts and rules. Here, facts have a different veracity than rules, as they are usually provided by a given (non-negotiable) case input and have to be kept "as-is" \cite{graham:2007business}. In the scope of inconsistency resolution, we are therefore only interested in identifying blamable rules, as these should be considered for re-modelling. Consequently, an element-based assessment for rule bases should only assign a payoff to blamable rules, and not facts. Correspondingly, this has to be considered when applying Shapley's game theoretical approach to distribute a blame mass over all elements. As recently discussed in \cite{corea2020towards}, the Shapley inconsistency value can accordingly be adjusted as follows. For that, let $\Free(B)$ denote the free formula in a rule base $B$, i.e., all $r\in B: r\not\in M, \forall M\in\MI(B)$.

\begin{definition}[Adjusted Shapley Inconsistency Value \cite{corea2020towards}]\label{def:adjustedShapley}
    Let $\inc$ be a rule-based inconsistency measure, $\rb$ be a rule base and $\alpha \in \rb$. Then, the adjusted Shapley inconsistency value of $\alpha$ w.r.t. $\inc$, denoted $S*_\alpha^{\inc}$ is defined via
    \begin{align*}
        &S*_\alpha^{\inc}(\rb)
        =
            \begin{cases}
                0 & \text{if } \alpha \in \F(\rb) \\
                \sum\limits_{B\subseteq\rb}( \mathit{CoalitionPayoff}_{\alpha,\rb}^{\inc}(B) +  \mathit{AdditionalPayoff}_{\alpha,\rb}^{\inc}(B) ) & \text{otherwise}
            \end{cases}\\
        \intertext{with}    
        &\mathit{CoalitionPayoff}_{\alpha,\rb}^{\inc}(B) = \frac{(b-1)!(n-b)!}{n!}(\inc(B)-\inc(B\setminus \alpha))\\
        \intertext{being the payoff for an element for any coalition $B\subseteq\rb$, and} 
        &\mathit{AdditionalPayoff}_{r,\rb}^{\inc}(B)
        = \left\{\begin{array}{ll}
            0& \text{if } r \in \Free(B)\\
            \frac{\sum_{f\in\F(B)}\mathit{CoalitionPayoff}_{f,\rb}^{\inc}(B)}{|r'\in \R(B) \text{ s.t. } r' \notin \Free(B)|} & \text{otherwise}
        \end{array}\right.
        \intertext{being the additional payoff that blamable rules receive, by shifting the blame mass from (given) facts to blamable rules.}
    \end{align*}
\end{definition}
\vspace{-1.2cm}
\begin{example}
Consider the rule bases $\rb_{2} = \{a, a \rightarrow b, a \rightarrow \neg b\}$ and $\rb_{3}  = \{a, a \rightarrow b, \neg b\}$. Then for the \emph{adjusted} Shapley inconsistency values w.r.t. $\incmi$, we have that $S*_a^{\incmi}(\rb_2) = 0, S*_{a\rightarrow b}^{\incmi}(\rb_2) = \frac{1}{3}  (+\frac{1}{3}/2) = \frac{1}{2}$, and $S*_{a\rightarrow \neg b}^{\incmi}(\rb_2) = \frac{1}{3}  (+\frac{1}{3}/2) = \frac{1}{2}$.  Also, we have that $S*_a^{\incmi}(\rb_3) = 0, S*_{\neg b}^{\incmi}(\rb_3) = 0$, and $S*_{a\rightarrow \neg b}^{\incmi}(\rb_3) = \frac{1}{3}  (+\frac{2}{3}) = 1$ . For the first rule base $\rb_2$ (containing two rules), the blame is evenly distributed amongst both rules. For the second rule base $\rb_3$, the single rule receives the entire blame. This assessment makes sense in a business rule setting, as the given fact input is evaluated against a set of humanly modelled rules, and any inconsistencies arise due to modelling errors in the set of business rules. The adjusted Shapley inconsistency values can thus be used for pin-pointing problematic rules for re-modelling purposes. 
\end{example}   

This element-based measure can consequently also be used to identify problematic rules over a multiset of cases, by inducing the corresponding multi-rb measure $m^\Sigma_{S*^I}$ (cf. Example \ref{ex:concludingExample}). As mentioned, an advantage of using the adjusted Shapley measure is that arbitrary inconsistency measures can be used to derive an element-based assessment over a set of cases, based on company needs. Furthermore, we can identify the following properties for an adjusted Shapley value $m^\Sigma_{S*^I}$. For this,
we assume that any inconsistency measure $\inc$ used to derive adjusted Shapley inconsistency values satisfies the basic properties of consistency', monotony' and free formula independence' as defined in \cite{hunter2010measure}\footnote{Let a rule base $\rb$ and an inconsistency measure $\inc$. Consistency' states that $\inc(\rb)=0$ iff $\rb$ is consistent. Monotony' states that if $\rb\subseteq \rb'$ then $\inc(\rb)\leq \inc(\rb')$. Free formula independence' states that If $\alpha\in\Free(\rb)$ then $\inc(\rb)=\inc(\rb\setminus\{\alpha\})$.}. 

\begin{proposition}
The adjusted multi-rb shapley inconsistency value satisfies \textsf{CO}, \textsf{MO} and \textsf{IN}.
\end{proposition}

Also, regarding the relation of inconsisteny measures and the corresponding $\Sigma-$induced Shapley inconsistency values for multi-rb analysis, we propose the following postulates.
\begin{description}
	\item[\emph{Distribution} (\textsf{DIS})]$\sum_{\alpha\in\R(\ms)}m^\Sigma_{S*^I}(\ms,\alpha) = m^\Sigma_I(\ms)$
	\item[\emph{Upper Bound} (\textsf{UB})]
	$\hat{V}^{S*^I}(\ms) \leq m^\Sigma_I(\ms)$
\end{description}
The first postulate states that the sum adjusted multi-rb Shapley inconsistency values over all rules is equal to the overall blame mass of the original multi-rb inconsistency measure $\inc$ (used as a parameter to derive the corresponding Shapley values). Also, the second property states that the adjusted multi-rb Shapley inconsistency values for an individual element cannot be greater than the overall assessment of the original multi-rb inconsistency measure $\inc$.

As we are only interested in identifying problematic rules (e.g. for re-modelling), it would also be plausible to adapt the property of fact minimality as proposed in \cite{corea2020towards} for a multi-rb use-case.
\begin{description}
	\item[\emph{Fact Minimality} (\textsf{FM})]$m^\Sigma_{S^I}(\ms,\alpha) = 0 \forall \alpha\notin\R(\ms)$.
\end{description}
This property states that (non-negotiable) facts should not be assigned any blame value in an element-based multi-rb assessment.

\begin{proposition}
The adjusted multi-rb shapley inconsistency value satisfies \textsf{DIS}, \textsf{UB} and \textsf{FM}.
\end{proposition}

We conclude with an example illustrating the introduced multi-rb measures.
\begin{example}\label{ex:concludingExample}We recall the set of business rule bases $\ms_1$ from Figure \ref{fig:exemplaryInstances} and its shared rule set $\R(\ms_1)=\{r_1,...,r_6\}$. A multi-rb assessment w.r.t. the introduced measures is then as follows.
\begin{align*}
    r_1: a &\rightarrow b &          m_{\culp_D}^{\Sigma}(\ms_1, r_1) &= 4 & m_{\culp_{\#}}^{\Sigma}(\ms_1, r_1)&= 5 & m_{S*^{\incmi}}^{\Sigma}(\ms_1, r_1)&= 2\\
    r_2: c &\rightarrow \neg b &          m_{\culp_D}^{\Sigma}(\ms_1, r_2) &= 3 & m_{\culp_{\#}}^{\Sigma}(\ms_1, r_2)&= 3 & m_{S*^{\incmi}}^{\Sigma}(\ms_1, r_2)&= 1.5\\
    r_3: b &\rightarrow x &          m_{\culp_D}^{\Sigma}(\ms_1, r_3) &= 2 & m_{\culp_{\#}}^{\Sigma}(\ms_1, r_3)&= 2 & m_{S*^{\incmi}}^{\Sigma}(\ms_1, r_3)&= 0.5\\
    r_4: x &\rightarrow z &          m_{\culp_D}^{\Sigma}(\ms_1, r_4) &= 2 & m_{\culp_{\#}}^{\Sigma}(\ms_1, r_4)&= 2 & m_{S*^{\incmi}}^{\Sigma}(\ms_1, r_4)&= 0.5\\
    r_5: y &\rightarrow \neg z &          m_{\culp_D}^{\Sigma}(\ms_1, r_5) &= 2 & m_{\culp_{\#}}^{\Sigma}(\ms_1, r_5)&= 2 & m_{S*^{\incmi}}^{\Sigma}(\ms_1, r_5)&= 0.5\\
    \cline{1-8}
    & & & & &&  m_{\incmi}^{\Sigma}(\ms_1)&= 5
    \end{align*}
    As can be seen, rule $r_1$ is classified as most problematic my all measures. This makes sense, as this rule is a cause of inconsistency over all cases for $\ms_1$. Hence, this rule should be prioritized in the scope of re-modelling and improving the set of business rules. Here, our proposed approach of multi-case inconsistency measurement can support modellers in identifying highly problematic rules from a global perspective, by recommending an order in which rules should be attended to, e.g. $<r1,r2,r3,r4,r5>$ for the shown example. A further important aspect for an application in practice is that the proposed measures can be combined to obtain multivariate metrics for a more fine-grained analysis. For example, as the $m_{\culp_D}^{\Sigma}(r)$ is equivalent to the number of distinct cases in which a rule $r$ is part of an inconsistency, this measure can be used to normalize and explain other measures. For example, a normalization via $m_{\culp_D}^{\Sigma}$ can be used to explain if a high $m_{\culp_{\#}}^{\Sigma}$-value originates from a few highly inconsistent cases (which might be outliers) or if the corresponding rule contributes a smaller amount towards inconsistency but in a vast majority of cases (in which case it might be sensible to re-consider this rule).
\end{example}

In this section, we have shown how arbitrary culpability measures can be transformed into multi-rb measures. Also, we have shown that the proposed measures satisfy desirable properties. Our results are summarized in Table \ref{tab:results}\footnote{Proofs can be found in the supplementary document (\url{https://bit.ly/2V2sIDw})}. 

\begin{table}[H]
	\footnotesize
	\parbox{1\linewidth}{
		\centering
		\begin{tabular}{|l|c|c|c|c|c|c|c|c|}
			\hline
			$\culp^m$      & \textsf{RS} & \textsf{RM} & \textsf{CO} & \textsf{MO} & \textsf{IN} & \textsf{DIS} & \textsf{UB} & \textsf{FM}\\
			\hline
			$m^\Sigma_{\culp_D}$ & \cmark      & \cmark      & \cmark   & \cmark  & \cmark  & \scriptsize{n/a}  & \scriptsize{n/a}   & \xmark     \\
			$m^\Sigma_{\culp_{\#}}$ &\cmark      & \cmark      & \cmark   & \cmark  & \cmark  & \scriptsize{n/a}  & \scriptsize{n/a}  & \xmark     \\
			$m^\Sigma_{S^{I}}$ & \cmark      & \cmark$^{c,i}$      & \cmark$^{c,i}$   & \cmark$^{m}$  & \cmark$^{i}$  & \cmark  & \cmark  & \cmark     \\
			\hline
		\end{tabular}
        \parbox{.44\textwidth}{
        \vspace{.2cm}
        
        \scriptsize{
        \emph{c:} \hspace{.1cm}If $\inc$ satisfies consistency'\\
        \emph{m:} If $\inc$ satisfies monotony'\\
        \emph{i:} \hspace{.15cm}If $\inc$ satisfies free formula independence'\\
        }
        }
		\caption{Compliance with rationality postulates of the investigated measures.}
		\label{tab:results}
	}
\end{table}

%
%
\section{Tool Support and Evaluation}\label{sec:evaluation}

We implemented our approach to assess element-based inconsistency over sequences of business rule cases\footnote{\url{https://gitlab.uni-koblenz.de/fg-bks/multi-rb-inconsistency-measurement/}}. Our implementation takes as input a shared business rule base and a sequence of fact sets, and can then computes the most problematic rules w.r.t. the multiset of rule bases. The $m^\Sigma_{\culp_D}$ and $m^\Sigma_{\culp_{\#}}$ measures can be used out-of-the-box, however, arbitrary culpability measures can be added based on company needs. To evaluate our tool, we then performed run-time experiments and investigated the computational complexity regarding various aspects of measuring inconsistency over sequences of business rule cases.

\subsection{Run-Time Experiments}

In the following, we present the results of run-time experiments with real-life and synthetic data-sets.

\subsubsection{Evaluation with real-life data sets.}
To evaluate the feasibility of applying our approach in practice, we conducted run-time experiments with real-life data sets of the \emph{Business Process Intelligence (BPI) challenge}\footnote{\url{https://data.4tu.nl/search?q=bpi+challenge}}. This yearly scientific challenge from the field of process management provides real-life process logs for evaluating approaches in an industrial setting. In a nutshell, we mined a rule set from each event log and then measured inconsistency over all cases of the respective log. Here, we analyzed the data-sets from the last four years, i.e., BPI'17 (log of a loan application process with 31,509 cases), BPI'18 (log of a fund distribution process with 43,809 cases), BPI'19 (log of an application process with 251,734 cases), and BPI'20 (log of a travel expense claim process with 10,500 cases). From these event logs, declarative constraints of the general business rule form in (\ref{eq:rule}) can be mined using the results from \cite{diciccio:2017resolving}. In this way, we were able to mine a rule set from each of the provided data sets. The resulting number of rules for the respective rule sets is provided in Table \ref{tab:runtime_results}. Also, the mined rule sets can be found online\footnote{\url{https://bit.ly/365Vs4C}}. We refer the reader to \cite{diciccio:2017resolving} for further details on the mining technique. Then, for each data set, we analyzed all cases as follows: 

    
    
    

For each data set, a shared rule base $R$ was mined as described above. Then, for all cases $C_1,...,C_n$, the individual case-dependent fact inputs $F_1,...,F_n$ were extracted from the log. We then constructed a multiset of rule bases $B_1,...,B_n$, where every $B_i=(R,F_i)$. We then applied our implementation to analyze inconsistencies over $B_1,...,B_n$ and measured the run-time. The results of our experiments are shown in Table \ref{tab:runtime_results}. The experiments were run on a machine with 3 GHz Intel Core i7 processor, 16 GB RAM (DDR3) under macOS.

\begin{table}[H]
    \centering
    \scriptsize{
    \begin{tabular}{p{1.6cm}p{1.6cm}p{1.6cm}p{1.6cm}p{3.2cm}}
    \hline
        Dataset & \# of Rules &\# of Cases & Runtime & \# of inconsistent Cases \\
    \hline
    BPI'17 & 50 & 31.509  & 5657s & 31.509 (100\%) \\
    BPI'18 & 84 & 43.809  & 3967s & 0 (0\%) \\
    BPI'19 & 51 & 251.734 & 1610s & 434 (0.17\%)  \\
    BPI'20 & 330 &10.500  & 1329s & 323 (3.07\%) \\
        &&&&\\
    \end{tabular}
    }
    \caption{Runtimes for analyzing all cases for the considered BPI data sets}
    \label{tab:runtime_results}
\end{table}
\vspace{-1cm}
As can be seen, an analysis of all cases was feasible for all data sets. 

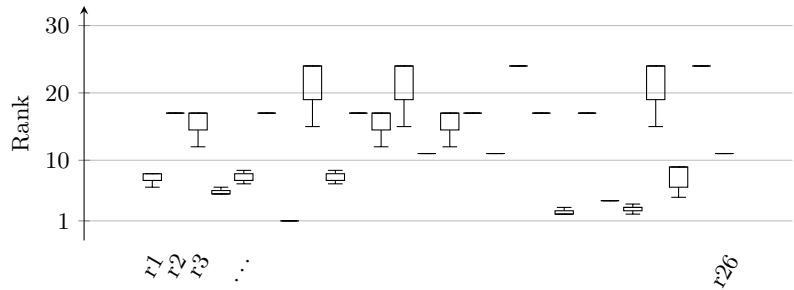
\begin{figure}[H]
    \centering

    \begin{tikzpicture}
	\pgfplotstableread[col sep=comma]{boxPlotData/bpi17.txt}\csvdata
	\pgfplotstabletranspose\datatransposed{\csvdata} 
	\begin{axis}[
		boxplot/draw direction = y,
		height=4.7cm,
		width=11cm,
		x axis line style = {opacity=0},
		axis x line* = bottom,
		axis y line = left,
		enlarge y limits,
		ymajorgrids,
		xtick = {1, 2, 3, 4,5,6,7,8,9,10,11,12,13,14,15,16,17,18,19,20,21,22,23,24,25,26},
		xticklabel style = {align=center, font=\small, rotate=60},
		xticklabels = {r1,r2,r3,,$\cdots$,,,,,,,,,,,,,,,,,,,,,r26},
		xtick style = {draw=none}, 
		ylabel = {Rank},
		ytick = {1,10,20,30},
		ymax=30
	]
		\foreach \n in {1,...,26} {
			\addplot+[boxplot, draw=black] table[y index=\n] {\datatransposed};
		}
	\end{axis}
    \end{tikzpicture}
    
    \caption{Rank distribution for the individual rules of the BPI'17 rule set over all cases.}
    \label{fig:boxplot}
\end{figure}

A central assumption of our approach is that a global perspective over all cases should be considered as opposed to viewing cases individually. Interestingly, this was also confirmed by our experiments with the above real-life data sets: 

For every individual rule base instance, we computed the $\culp_{\#}$ values for all rules and then ranked all rules by this value (rank 1 meaning that this rule is the most problematic element, and so on). If $n$ rules had the same $\culp_{\#}$ value, they were assigned the sum of the occupied ranks divided by $n$ (e.g. if two rules had the highest $\culp_{\#}$ value, they were awarded the rank (1+2)/2 = 1.5, and the next rule had the rank 3). Figure \ref{fig:boxplot} shows the distribution of all assigned ranks for the rules for the BPI'17 data set over all cases. For readability, rules that did not participate in any inconsistencies are omitted.

While there were some rules that had the same rank in all cases (e.g. r2), there were many rules where the respective local rankings had a large variability (e.g. r3). This shows that the global perspective as proposed in this work should be strongly considered in the scope of auditing. 

In general, we see the above experiments as positive in regard to applying our approach in practice. As the analyzed data-sets were unrelated, no further comparison of run-times can be made. Therefore, we further assess our approach with synthetic data sets.

\subsubsection{Evaluation with synthetic data sets.}
We created a generator for synthetic rule base instances. Our generator can produce set of business rule base instances, based on a shared rule set and a sequence of fact sets relative to this rule set. As parameters, our generator takes the desired rule base size and a desired number of cases. Then, the generator constructs a multiset of rule bases as follows:

The set of business rule instances is constructed over a (potentially infinite) alphabet $\mathfrak{A}=<a,b,...>$. Then, for a desired number of rules $n_r$, a rule set $R=\{r_1,...,r_{n_r}\}$ is generated, where every $r_i$ is of the form $\mathfrak{A}_i \rightarrow \neg\mathfrak{A}_{i+1}$. For example, for a parameter of $2$ desired rules, the resulting rule set is $R=\{a\rightarrow\neg b, b\rightarrow \neg c\}$. To then generate a desired number of random cases $n_c$, a multiset of fact sets $F=\{F_1,...,F_{n_c}\}$ is initialized. Then, each of the fact sets is populated by adding atoms of the rule base, based on a user-defined probability. For example, for the exemplary rule set $R$ of size $2$, a random fact set relative to $R$ could be any element of $\{\emptyset,a,b,c,ab,ac,bc,abc\}$.  In this way, the generator can create a set of random rule base instances $B=\{B_1,...,B_{n_c}\}$, where every $B_i=(R,F_i)$. 

\begin{figure}
 \centering
 \pgfplotsset{scaled y ticks=false}
 \resizebox{.7\textwidth}{!}{%
 \begin{tikzpicture}
     \begin{axis}[xlabel=Rule base size, ylabel=Number of cases, zlabel=Run-time in seconds, xlabel style={sloped like x axis}, ylabel style={sloped},yticklabels={1,2,50T,100T,5,6,7,8,9,10}, view/az=35, zmin=0, grid=major]
         \addplot3[surf, shader=faceted]
         coordinates {
(10,10000,89.7)
(10,20000,107.8)
(10,30000,126.2)
(10,40000,174.0)
(10,50000,176.7)
(10,60000,224.3)
(10,70000,224.8)
(10,80000,251.3)
(10,90000,292.9)
(10,100000,403.1)

(20,10000,81.7)
(20,20000,146.5)
(20,30000,219.2)
(20,40000,299.9)
(20,50000,350.8)
(20,60000,393.1)
(20,70000,441.4)
(20,80000,544.5)
(20,90000,696.1)
(20,100000,976.3)

(30,10000,158.8)
(30,20000,312.7)
(30,30000,397.6)
(30,40000,447.8)
(30,50000,561.2)
(30,60000,653.2)
(30,70000,914.5)
(30,80000,861.4)
(30,90000,969.0)
(30,100000,1133.6)

(40,10000,153.6)
(40,20000,256.5)
(40,30000,401.9)
(40,40000,626.3)
(40,50000,731.5)
(40,60000,1042.7)
(40,70000,1147.8)
(40,80000,1091.3)
(40,90000,1256.0)
(40,100000,1282.0)

(50,10000,155.6)
(50,20000,310.3)
(50,30000,500.9)
(50,40000,871.5)
(50,50000,1017.0)
(50,60000,1175.2)
(50,70000,1417.2)
(50,80000,1581.7)
(50,90000,1752.2)
(50,100000,2116.0)

(60,10000,279.1)
(60,20000,418.6)
(60,30000,790.0)
(60,40000,941.5)
(60,50000,1225.4)
(60,60000,1296.3)
(60,70000,1756.4)
(60,80000,1848.8)
(60,90000,2010.9)
(60,100000,2402.9)

(70,10000,296.2)
(70,20000,463.4)
(70,30000,944.7)
(70,40000,1026.9)
(70,50000,1323.3)
(70,60000,1802.6)
(70,70000,2097.8)
(70,80000,2181.9)
(70,90000,2482.4)
(70,100000,2804.1)

(80,10000,309.9)
(80,20000,565.3)
(80,30000,909.8)
(80,40000,1377.8)
(80,50000,1746.1)
(80,60000,1922.9)
(80,70000,2246.1)
(80,80000,2560.4)
(80,90000,2656.5)
(80,100000,3210.3)

(90,10000,308.4)
(90,20000,775.6)
(90,30000,1096.7)
(90,40000,1280.2)
(90,50000,1612.4)
(90,60000,2126.2)
(90,70000,2561.9)
(90,80000,3187.5)
(90,90000,2959.6)
(90,100000,3801.2)

(100,10000,400.1)
(100,20000,721.2)
(100,30000,1143.7)
(100,40000,1709.8)
(100,50000,1561.0)
(100,60000,1864.3)
(100,70000,2178.7)
(100,80000,2516.8)
(100,90000,2912.1)
(100,100000,3111.9)
       };
  \end{axis}
 \end{tikzpicture}
 }
\caption{Run-times for the analysis of 10x10 synthetic sets of rule base instances} 
\label{fig:surface}
\end{figure}
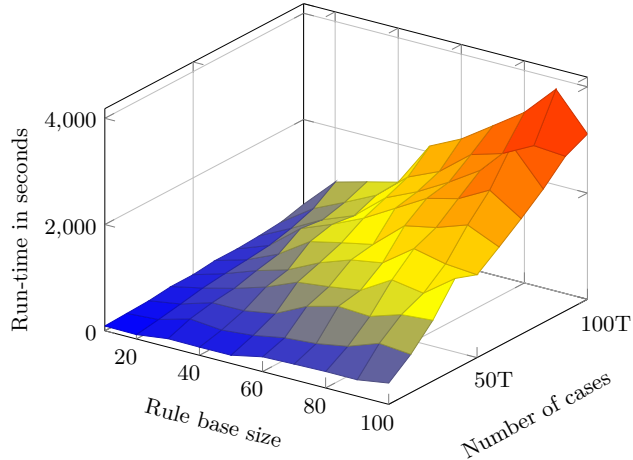

An advantage of our generator is that the structure of the contained rules is similar in all rule bases, which thus allows for better comparability. We consequently used our generator to analyze multiple sets of business rule cases with different parameters and measured the run-times. As parameter settings, based on the observed sizes from the real-life data-sets, we selected as parameters the rule base size from 10,20,...,100, and the number of cases from 10.000,20.000,...,100.000 and then tested every possible combination. Thus, a total of 100 (10x10) different configurations were tested (see above for hardware). The results of our experiments are shown in Figure \ref{fig:surface}. The smallest setting (10 rules and 10.000 cases) took around 90s, where the largest configuration (100 rules and 100.000 cases) took around 50 minutes. As can be seen, the run-time scales proportionally with the size of the rule base and the number of cases. Thus, we could not identify any of these two factors to be a dominant limiting factor to the run-times in our experiments.

To summarize, both the evaluation with real-life data sets and with synthetic data sets yielded feasible run-times. To extend this empirical analysis, we continue with an investigation of computational complexity in regard to our proposed approach.

\subsection{Complexity Analysis\protect\footnote{Proofs can be found in the supplementary document (\url{https://bit.ly/2V2sIDw})}}
We assume familiarity with basic concepts of computational complexity and basic complexity classes such as $\textsf{P}$ and $\textsf{NP}$, see \cite{Papadimitriou:1994a} for an introduction. We first observe that the satisfiability problem for our formalism of business rules bases is tractable (note that similar observations have been made before on similar formalisms, see e.\,g., \cite{Dantsin:1997}).
\begin{proposition}\label{prop:sat}
    Let $\rb$ be a rule base. The problem of deciding whether $\rb$ is consistent can be solved in polynomial time.
\end{proposition}
Then, the complexity of deciding whether a certain rule is contributing to the overall inconsistency is as follows.
\begin{proposition}
    Let $\ms$ be a multiset of rule bases with $\ms=(\rb_1, ..., \rb_n)=(\{\F_1\cup\R\},...,\{\F_n\cup\R\})$ and let $r\in \rb$. The problem of deciding whether there is a $i\in\{1,\ldots,n\}$ and $M\in\MI(\rb_i)$ s.\,t. $r\in M$ is \textsf{NP}-complete.
\end{proposition}

The following two results deal with the computational complexity of computing the baseline measure $\culp_{\#}$.
\begin{proposition}\label{prop:mi}
    Let $\rb$ be a rule base and $M\subseteq \rb$. The problem of deciding whether $M\in\MI(\rb)$ can be solved in polynomial time.
\end{proposition}

For our final result note that $\#\textsf{P}$ is the complexity class of counting problems where the problem of deciding whether a particular element has to be counted is in $\textsf{P}$, cf.\ \cite{Valiant:1979}. 
\begin{proposition}
    Let $\ms$ be a multiset of rule bases with $\ms=(\rb_1, ..., \rb_n)=(\{\F_1\cup\R\},...,\{\F_n\cup\R\})$ and let $r\in \R$. The problem of determining $|\{M \in \MI(\rb_i) \mid i\in \{1,\ldots,n\} r \in M\}|$ is $\#\textsf{P}$-complete.
\end{proposition}
%
%
\section{Conclusion}\label{sec:conclusion}
In this work, we have shown how arbitrary culpability measures (for single rule bases) can be automatically transformed into multi-rb measures while maintaining desirable properties. This is highly needed in practice, as companies are often faced with thousands of rule bases daily, and thus need means to assess inconsistency from a global perspective. Here, our proposed measures can be used to gain fine-grained insights into inconsistencies in sequences of business rule bases. For the analyzed (real-life) data-sets, the proposed multi-cases analysis could be performed in a feasible run-time. Intuitively, the number of cases or the size of the rule base affect the run-time of our approach. Here, we plan to develop more efficient algorithms in future work. As a main takeaway, our results indicate that the interrelations of individual cases need to be considered for business rules management, which should be addressed more in future works.

\bibliographystyle{splncs04}
\bibliography{references}

\begin{thebibliography}{10}
\providecommand{\url}[1]{\texttt{#1}}
\providecommand{\urlprefix}{URL }
\providecommand{\doi}[1]{https://doi.org/#1}

\bibitem{corea:2019b}
Corea, C., Deisen, M., Delfmann, P.: Resolving inconsistencies in declarative
  process models based on culpability measurement. In: 15. Internationale
  Tagung Wirtschaftsinformatik, {WI} 2019 (2019)

\bibitem{Corea:2020a}
Corea, C., Thimm, M.: On quasi-inconsistency and its complexity. AI
  \textbf{284} (2020)

\bibitem{corea2020towards}
Corea, C., Thimm, M.: Towards inconsistency measurement in business rule bases.
  In: Proceedings of the 24th European Conference on Artificial Intelligence
  (ECAI 2020), Santiago de Compostela, Spain, 2020 (2020)

\bibitem{Dantsin:1997}
Dantsin, E., Eiter, T., Gottlob, G., Voronkov, A.: Complexity and expressive
  power of logic programming. In: Proceedings of the 12th Annual IEEE
  Conference on Computational Complexity (CCC'97). pp. 82--101 (1997)

\bibitem{diciccio:2017resolving}
Di~Ciccio, C., Maggi, F.M., Montali, M., Mendling, J.: Resolving
  inconsistencies and redundancies in declarative process models. Inf. Systems
  \textbf{64},  425--446 (2017)

\bibitem{graham:2007business}
Graham, I.: Business rules management and service oriented architecture: a
  pattern language. John wiley \& sons (2007)

\bibitem{Grant:2018}
Grant, J., Martinez, M.V. (eds.): Measuring Inconsistency in Information.
  College Publications (2018)

\bibitem{hunter2010measure}
Hunter, A., Konieczny, S.: On the measure of conflicts: Shapley inconsistency
  values. Artificial Intelligence  \textbf{174}(14),  1007--1026 (2010)

\bibitem{hunter2008measuring}
Hunter, A., Konieczny, S., et~al.: Measuring inconsistency through minimal
  inconsistent sets. KR  \textbf{8},  358--366 (2008)

\bibitem{mcareavey2014computational}
McAreavey, K., Liu, W., Miller, P.: Computational approaches to finding and
  measuring inconsistency in arbitrary knowledge bases. International Journal
  of Approximate Reasoning  \textbf{55}(8),  1659--1693 (2014)

\bibitem{Papadimitriou:1994a}
Papadimitriou, C.: Computational Complexity. Addison-Wesley (1994)

\bibitem{potyka2018measuring}
Potyka, N.: Measuring disagreement among knowledge bases. In: International
  Conference on Scalable Uncertainty Management. pp. 212--227. Springer (2018)

\bibitem{thimm:2017compliance}
Thimm, M.: On the compliance of rationality postulates for inconsistency
  measures: A more or less complete picture. KI  \textbf{31}(1),  31--39 (2017)

\bibitem{Thimm:2019d}
Thimm, M.: Inconsistency measurement. In: Proceedings of the 13th International
  Conference on Scalable Uncertainty Management (SUM'19) (2019)

\bibitem{Valiant:1979}
Valiant, L.: The complexity of computing the permanent. Theoretical Computer
  Science  \textbf{8},  189--201 (1979)

\end{thebibliography}

\newpage
\section*{Appendix A: Proofs of Technical Results}
\setcounter{proposition}{0}
\begin{proposition}
$m_{\culp_D}^{\Sigma}$ and $m_{\culp_{\#}}^{\Sigma}$ satisfy \textsf{RS} and \textsf{RM}.
\begin{proof}
We consider $m_{\culp_D}^{\Sigma}$ and $m_{\culp_{\#}}^{\Sigma}$ in turn. For this, let $\ms$ be a multiset of rule bases, $B$ any rule base in $\ms$, and $r$ any rule in a rule base $B$.
\begin{itemize}
    \item We start with the measure $m_{\culp_D}^{\Sigma}$. To show \textbf{\textsf{rule symmetry}}, as $m_{\culp_D}^{\Sigma}(\ms,r)=\sum_ {B\in\ms}\culp_D(B,r)$, we have that $m_{\culp_D}^{\Sigma}(\ms,r)=m_{\culp_D}^{\Sigma}((\rb_{1},...\rb_n),r)$, for any permutation of the order of $\rb_1$ to $\rb_n$ due to commutativity via $\sum_{B_i\in\ms}\culp_D(B_i,r) = \culp_D(B_1,r)+ ... + \culp_D(B_n,r)$, with $n=|\{B\in\ms\}|$. For \textbf{\textsf{rule minimality}}, recall that a rule $r$ is defined as a free formula in $\ms$ if $r \notin M, \forall M \in \bigcup_{b\in\ms}\MI(b)$. Consequently, if $r\in\Free(\ms)$, then $\sum_ {B\in\ms}\culp_D(B,r) = 0$ per definition.
    \item The proofs for $m_{\culp_{\#}}^{\Sigma}$ are analogous, i.e., $m_{\culp_{\#}}^{\Sigma}(\ms,r)=m_{\culp_{\#}}^{\Sigma}((\rb_{1},...\rb_n),r)$, for any permutation of the order of $\rb_1$ to $\rb_n$ due to commutativity, and $\sum_ {B\in\ms}\culp_{\#}(B,r) = 0$ if $r\in\Free(\ms)$, as $\culp_{\#}(B,r) = |\{M \in \MI(B) \mid r \in M\}|$ for any $\rb\in\ms$.
\end{itemize}
\end{proof}
\end{proposition}

\setcounter{proposition}{1}
\begin{proposition}\label{prop:rs}
Any $\Sigma$-induced multi-rb culpability measure satisfies \textsf{RS}. Given a culpability measure $\culp$ satisfying \textsf{MIN}, any multi-rb culpability measure $\Sigma$-induced via $\culp$ satisfies \textsf{RM}.
\begin{proof}
Let $\ms$ be a multiset of rule bases, $B$ any rule base in $\ms$, and $r$ any rule in a rule base $B$. To show \textbf{\textsf{rule symmetry}}, as $m_{\culp^m}^{\Sigma}(\ms,r)=\sum_ {B\in\ms}\culp^m(B,r)$, $\culp^m(\ms,r)=\culp^m((\rb_{1},...\rb_n),r)$ for any permutation of the order of $\rb_1$ to $\rb_n$ due to commutativity via $\sum_{B_i\in\ms}\culp^m(B_i,r) = \culp^m(B_1,r)+ ... + \culp^m(B_n,r)$, with $n=|\{B\in\ms\}|$. To show \textbf{\textsf{rule minimality}}, given a culpability measure $\culp$, if we have for a rule $r\in B$: $r\not\in M, \forall M \in \MI(B)$ and $\culp(B,r)=0$ , then $\culp^m = \sum_ {B\in\ms}\culp(B,r) = 0$ per assumption.
\end{proof}
\end{proposition}

\setcounter{proposition}{2}
\begin{proposition}
$m_{\culp_D}^{\Sigma}$ and $m_{\culp_{\#}}^{\Sigma}$ satisfy \textsf{CO}, \textsf{MO} and \textsf{IN}.
\begin{proof}
We consider $m_{\culp_D}^{\Sigma}$ and $m_{\culp_{\#}}^{\Sigma}$ in turn. For this, let $\ms$ be a multiset of rule bases, $B$ any rule base in $\ms$, and $r$ any rule in a rule base $B$.
\begin{itemize}
    \item We start with the measure $m_{\culp_D}^{\Sigma}$. To show \textbf{\textsf{multiset consistency}}, observe that $\culp_D(X,r)$ for a consistent rule base $X$ is $0$ for any rule $r$ per definition, as $\MI(X)=\emptyset$. Thus, if $\nexists B\in\ms:B\models\perp$, then $m_{\culp_D}^{\Sigma}(\ms)=0$. In turn, $\hat{V}^{m_{\culp_D}^{\Sigma}}(\ms)=\mathit{max}_{r\in\R(\ms)}(\culp_D(\ms,r))=0$. For the other direction, assume that $\hat{V}^{m_{\culp_D}^{\Sigma}}(\ms)>0$ for a case where $\nexists \rb\in\ms:\rb\models\perp$. This would mean, that there must exist a rule $r$ in a consistent rule base $X$ of $\ms$, s.t. $\culp_D(X,r)\neq 0$. This contradicts $\culp_D$ by definition. To show \textbf{\textsf{multiset monotony}}, observe that if a rule $r'$ is added, for any rule base $B_i$, we have that $\culp_D(B_i,r')= 0$ or $1$. Thus, $m_{\culp_D}^{\Sigma}(\ms\cup \{r'\})\geq m_{\culp_D}^{\Sigma}(\ms)$. Hence, $\hat{V}^{m_{\culp_D}^{\Sigma}}(\ms \cup \{r'\})=\mathit{max}_{r\in\R(\ms)}(\culp_D(\ms\cup \{r'\},r)) \geq \hat{V}^{m_{\culp_D}^{\Sigma}}(\ms)$. To show \textbf{\textsf{multiset free formula independence}}, recall that a rule $r'$ is defined as a free formula in $\ms$ if $r' \notin M, \forall M \in \bigcup_{b\in\ms}\MI(b)$. Consequently, if $r'\in \Free(\ms\cup \{r'\})$, then $|\MI(B_i\cup \{r'\})|=|\MI(B_i)|$ for all $B\in\ms$. In turn, for any other rule $r$ in a rule base $B_i\in\ms$, $\culp_D(B_i,r)=\culp_D(B_i\cup\{r'\},r)$. In result, if $r'\in \Free(\ms\cup\{r'\})$, then $\hat{V}^{m_{\culp_D}^{\Sigma}}(\ms \cup \{r'\}) = \hat{V}^{m_{\culp_D}^{\Sigma}}(\ms)$.
    \item The proofs for $m_{\culp_{\#}}^{\Sigma}$ are analogous, i.e., if $\nexists B\in\ms:B\models\perp$ then $m_{\culp_{\#}}^{\Sigma}(\ms)=0$ as $|\MI(B_i)|=0$ for any $B_i\in\ms$, $|\MI(B\cup \{r\})|\geq |\MI(B)|$ for any rule $r$, resp. $|\MI(B\cup \{r\})| = |\MI(B)|$ if $r$ is a free formula in $(B\cup \{r\})$.
\end{itemize}

\end{proof}
\end{proposition}

\setcounter{proposition}{3}
\begin{proposition}
The adjusted multi-rb shapley inconsistency value satisfies \textsf{CO}, \textsf{MO}, \textsf{IN}(, \textsf{RS} and \textsf{RM}).
\begin{proof}
Let $\ms$ be a multiset of rule bases, $B$ any rule base in $\ms$, and $r$ any rule in a rule base $B$. Also, recall that we assume any inconsistency measure $\inc$ that is used to derived an adjusted multi-rb shapley inconsistency value $m^\Sigma_{S*^{I}}$ satisfies consistency', monotony' and free formula independence'. We then consider the individual properties in turn. To show \textbf{\textsf{multiset consistency}}, observe that for a consistent rule base $B$, we have that $\inc(B)=0$ per assumption of consistency'. It follows that for a consistent rule base $B$, $S*_\alpha^{\inc}(B)=0$ for any $\alpha\in B$, thus $\hat{V}^{m^\Sigma_{S*^{I}}}(\ms)=\mathit{max}_{r\in\R(\ms)}(m^\Sigma_{S*^{I}}(B,r))=0$ if $\nexists \rb\in\ms:\rb\models\perp$. For the only if direction, assume we would have a rule $\alpha$ s.t. $S*_\alpha^{\inc}(B)\neq0$ for a consistent rule base $B$. This would mean $\inc(B)-\inc(B\setminus\alpha)>0$ for a consistent rule base $B$, which contradicts the assumption of consistency'. To show \textbf{\textsf{multiset monotony}}, observe that $\inc(B\cup \{r\}) \geq \inc(B)$ for any rule $r$ if $\inc$ satisfies monotony'. Therefore, $S*_\alpha^{\inc}(B\cup \{r\})\geq S*_\alpha^{\inc}(B)$ for any $\alpha\in B$. It follows that $m^\Sigma_{S*^I}(\ms \cup \{r\},\alpha)\geq m^\Sigma_{S*^I}(\ms,\alpha)$ for any rule $\alpha$, and thus $\hat{V}^{m^\Sigma_{S*^{I}}}(\ms\cup\{r\})\geq \hat{V}^{m^\Sigma_{S*^{I}}}(\ms)$. The proof for \textbf{\textsf{multiset free formula independence}} is analogous, i.e., $\inc(\MI(B\cup \{r\})) = \inc(\MI(B))$ for any rule $r\in\Free(B\cup\{r\})$ if $\inc$ satisfies free formula independence'. Next, \textbf{\textsf{rule symmetry}} follows from Proposition \ref{prop:rs}. Last, to show \textbf{\textsf{rule minimality}}, it suffices to show that $S*^{\inc}$ satisfies minimality, which has been shown in \cite{corea2020towards}, i.e., for any fact $f$, $S*_f^{\inc}=0$ per definition, and for any free rule $\alpha$,  $\inc(B)-\inc(B\setminus\alpha)$ in the last part of the summand of the coalition payoff will always equate to $0$ due to the consistency' and free formula independence' assumption of the underlying measure $\inc$, thus, for any $r_i\in\R(\rb)$, if $r_i\not\in M, \forall M \in \MI(\rb)$, then $S*_{r_i}^{\inc}=0$.
\end{proof}
\end{proposition}

\setcounter{proposition}{4}
\begin{proposition}
The adjusted multi-rb shapley inconsistency value satisfies \textsf{DIS}, \textsf{UB} and \textsf{FM}.
\begin{proof}
Let $\ms$ be a multiset of rule bases, $B$ any rule base in $\ms$, and $r$ any rule in a rule base $B$. We then consider the individual properties in turn. The proof for \textbf{\textsf{distribution}} follows the proof in \cite{corea2020towards}: We recall the adjusted Shapley inconsistency value
\begin{align*}
    \scriptsize{
    S*_\alpha^{\inc}(\rb)
    =
        \begin{cases}
            0 & \text{if } \alpha \in \F(\rb) \\
            \sum\limits_{B\subseteq\rb} \mathit{CoalitionPayoff}_{\alpha,\rb}^{\inc}(B) + 
            \sum\limits_{B\subseteq\rb}\mathit{AdditionalPayoff}_{\alpha,\rb}^{\inc}(B) & \text{otherwise}
        \end{cases}
        }
\end{align*}
In the following, we abbreviate \emph{CoalitionPayoff} as \emph{CP} and \emph{AdditionalPayoff} as \emph{AP} for readability. Then, for a set of rule bases $\ms$, we consider the sum of all adjusted Shapley values (for all elements in $\ms$ over all rule bases $\rb\in\ms$).
{\footnotesize
\begin{align*}
    &\sum_{\alpha \in \rb} \sum_ {B\in\ms} S*_\alpha^{\inc}(\rb)\\
    &=\sum_{\alpha \in \rb} \sum_ {B\in\ms}
        \begin{cases}
            0 & \text{if } \alpha \in \F(\rb) \\
            \sum\limits_{B\subseteq\rb} \mathit{CP}_{\alpha,\rb}^{\inc}(B) + 
            \sum\limits_{B\subseteq\rb}\mathit{AP}_\alpha^{\inc}(B) & \text{otherwise}
        \end{cases}\\
    &=\sum_{\alpha \in \rb}\sum_ {B\in\ms}\sum\limits_{B\subseteq\rb} \mathit{CP}_{\alpha,\rb}^{\inc}(B) - \sum_{f \in \F(\rb)}\sum_ {B\in\ms}\sum\limits_{B\subseteq\rb} \mathit{CP}_{f,\rb}^{\inc}(B)+\sum_{\alpha \in \R(\rb)}\sum_ {B\in\ms}\sum\limits_{B\subseteq\rb}\mathit{AP}_{\alpha,\rb}^{\inc}(B)\\
    \intertext{Following \cite{hunter2010measure}, the first summand can be rewritten.}
    &=\sum_ {B\in\ms}\inc(\rb) - \sum_{f \in \F(\rb)}\sum_ {B\in\ms}\sum\limits_{B\subseteq\rb} \mathit{CP}_{f,\rb}^{\inc}(B)+\sum_{\alpha \in \R(\rb)}\sum_ {B\in\ms}\sum\limits_{B\subseteq\rb}\mathit{AP}_{\alpha,\rb}^{\inc}(B)\\
    \intertext{Then}
    &=\sum_ {B\in\ms}\inc(\rb) - \sum_{f \in \F(\rb)}\sum_ {B\in\ms}\sum\limits_{B\subseteq\rb} \mathit{CP}_{f,\rb}^{\inc}(B)+\sum_{\alpha \in \R(\rb)}\sum_ {B\in\ms}\sum\limits_{B\subseteq\rb} \left\{\begin{array}{ll}
        0& \text{if } r \in \Free(B)\\
        \frac{\sum_{f\in\F(B)}\mathit{CP}_{f,\rb}^{\inc}(B)}{|r\in \R(B) \text{ s.t. } r \notin \Free(B)|} & \text{otherwise}
        \end{array}\right.  \\
    &=\sum_ {B\in\ms}\inc(\rb) - \sum_{f \in \F(\rb)}\sum_ {B\in\ms}\sum\limits_{B\subseteq\rb} \mathit{CP}_{f,\rb}^{\inc}(B)+\sum_{r \in \Free(\R(\rb))}\sum_ {B\in\ms}\sum\limits_{B\subseteq\rb} 0 + \sum_{r \not\in \Free(\R(\rb))}\sum_ {B\in\ms}\sum\limits_{B\subseteq\rb} \frac{\sum_{f\in\F(B)}\mathit{CP}_{f,\rb}^{\inc}(B)}{|r\in \R(B) \text{ s.t. } r \notin \Free(B)|}\\
    &=\sum_ {B\in\ms}\inc(\rb) - \sum_{f \in \F(\rb)}\sum_ {B\in\ms}\sum\limits_{B\subseteq\rb} \mathit{CP}_{f,\rb}^{\inc}(B) + \sum_{r \not\in \Free(\R(\rb))}\sum_ {B\in\ms}\sum\limits_{B\subseteq\rb} \frac{\sum_{f\in\F(B)}\mathit{CP}_{f,\rb}^{\inc}(B)}{|r\in \R(B) \text{ s.t. } r \notin \Free(B)|}\\
    &=\sum_ {B\in\ms}\inc(\rb) - \sum_{f \in \F(\rb)}\sum_ {B\in\ms}\sum\limits_{B\subseteq\rb} \mathit{CP}_{f,\rb}^{\inc}(B) + \sum\limits_{B\subseteq\rb}\sum_ {B\in\ms} \sum_{f\in\F(B)}\mathit{CP}_{f,\rb}^{\inc}(B)\\ 
    &=\sum_ {B\in\ms}\inc(\rb) - \sum_{f \in \F(\rb)}\sum_ {B\in\ms}\sum\limits_{B\subseteq\rb} \mathit{CP}_{f,\rb}^{\inc}(B) + \sum_{f\in\F(B)}\sum_ {B\in\ms}\sum\limits_{B\subseteq\rb}\mathit{CP}_{f,\rb}^{\inc}(B)\\
    &=\sum_ {B\in\ms}\inc(\rb)
\end{align*}
}
To show \textbf{\textsf{upper bound}}, observe that due to $\sum_{\alpha \in \rb} \sum_ {B\in\ms} S*_\alpha^{\inc}(\rb)=\sum_ {B\in\ms}\inc(\rb)$ via distribution, we have that $\hat{V}^{m_{S*^I}^{\Sigma}}(\ms)=\mathit{max}_{r\in\R(\ms)}(m_{S*^I}(\ms,r)) \leq m^\Sigma_I(\ms)$. Last, to show \textbf{\textsf{fact minimality}}, observe that $S*^I_f(B)=0$ for any fact $f\in B$ per definition, thus
$m^\Sigma_{S^I}(\ms,\alpha) = 0 \forall \alpha\notin\R(\ms)$.

\end{proof}
\end{proposition}

\setcounter{proposition}{5}
\begin{proposition}\label{prop:sat}
    Let $\rb$ be a rule base. The problem of deciding whether $\rb$ is consistent can be solved in polynomial time.
    \begin{proof}
        The minimal model $M$ of $\rb$ can be determined as follows:
            \begin{enumerate}
                \item $M=\F(\rb)$
                \item Let $r\in \R(\rb)$ be s.\,t. $body(r)\subseteq M$
                \item If there is no such rule, then return $M$
                \item Otherwise, $M:=M\cup \{head(r)\}$ and continue with 2.
            \end{enumerate}
            I can be seen that $M$ is both closed and minimal and therefore the minimal model of $\rb$. Both, the algorithm above and checking whether $M$ is inconsistent are polynomial, therefore deciding whether $\rb$ is consistent is polynomial.
    \end{proof}
\end{proposition}

\setcounter{proposition}{6}
\begin{proposition}
    Let $\ms$ be a multiset of rule bases with $\ms=(\rb_1, ..., \rb_n)=(\{\F_1\cup\R\},...,\{\F_n\cup\R\})$ and let $r\in \rb$. The problem of deciding whether there is a $i\in\{1,\ldots,n\}$ and $M\in\MI(\rb_i)$ s.\,t. $r\in M$ is \textsf{NP}-complete.
    \begin{proof}
        For \textsf{NP}-membership consider the following non-deterministic algorithm:
        \begin{enumerate}
            \item Guess $i\in\{1,\ldots,n\}$
            \item Guess a set $M\subseteq \rb_i$ with $r\in M$
            \item If $M$ is consistent, return \textsc{False}
            \item For each $x\in M$, if $M\setminus\{r\}$ is inconsistent return \textsc{False}
            \item Return \textsc{True}
        \end{enumerate}
        Observe that the above algorithm runs in polynomial non-deterministic time (due to consistency checks being polynomial, cf. Proposition~\ref{prop:sat}) and returns \textsc{True} iff $r$ is contained in a minimal inconsistent subset of at least one of $\rb_1, ..., \rb_n$.
        
        In order to show \textsf{NP}-hardness, we reduce the problem 3\textsc{Sat} to the above problem. For that, let $I=\{c_1,\ldots,c_n\}$ be a set of clauses $c_i=\{l_{i,1},l_{i,2},l_{i,2}\}$ where each $l_{i,j}$ is a literal of the form $a$ or $\neg a$ (with an atom a). 3\textsc{Sat} then asks whether there is an assignment $i:A\rightarrow \{\textsc{True},\textsc{False}\}$ that satisfies all clauses of $I$, where $A$ is the set of all atoms appearing in $I$. We introduce new atoms $k_1,\ldots,k_n$ for each of the clauses and a new atom $s$ (indicating satisfiability) and define $\rb_I$ through
        \begin{align*}
            \F(\rb_I) & = \{ a, \neg a\mid a\in A\} \cup\{\neg s\}\\
            \R(\rb_I) & = \{l_{i,j} \rightarrow k_i \mid j=1,2,3, i=1,\ldots, n\}\cup \{r=k_1,\ldots, k_n\rightarrow s\}
        \end{align*}
        We now claim that $I$ is satisfiable iff $r$ is in a minimal inconsistent subset of $\rb_I$ (which is a special case of our problem with $\ms=(\rb_I)$). So assume $I$ is satisfiable and let $i$ be a satisfying assignment. Observe that $M$ defined via
        \begin{align*}
            M  & = \{a \mid i(a)=\textsc{True} \} \cup \{\neg a \mid i(a)=\textsc{False}\}\cup\{\neg s\}\cup \R(\rb_I)
        \end{align*}
        is inconsistent: as $i$ is a satisfying assignment, each $k_i$ ($i=1,\ldots,n$) can be derived in $M$; then $s$ can also be derived, producing a conflict with $\neg s$. On the other hand, note that $M\subseteq \{r\}$ is consistent. It follows that there is a minimal inconsistent set $M'\subseteq M$ with $r\in M'$.
        
        Now assume that there is a minimal inconsistent set $M\subseteq \rb_I$ with $r\in M$. First observe that there is no atom $a$ s.\,t., $a,\neg a\in M$ (otherwise $M\setminus\{r\}$ would still be inconsistent). Let $i:A\rightarrow \{\textsc{True},\textsc{False}\}$ be any assignment with $i(a)=\textsc{True}$ if $a\in M$ and $i(a)=\textsc{False}$ if $\neg a\in M$. It follows that each clause $c_1,\ldots,c_n$ is satisfied by $i$ (as each $k_i$ could be derived in $M$) and so $i$ is a satisfying assignment for $I$.
    \end{proof}
\end{proposition}

\setcounter{proposition}{7}
\begin{proposition}\label{prop:mi}
    Let $\rb$ be a rule base and $M\subseteq \rb$. The problem of deciding whether $M\in\MI(\rb)$ can be solved in polynomial time.
    \begin{proof}
        Deciding whether $M$ is inconsistent and $M\setminus \{x\}$ for each $x\in \rb$ is consistent can each be solved in polynomial time due to Proposition~\ref{prop:sat}. It follows that deciding  $M\in\MI(\rb)$ can be solved in polynomial time.
    \end{proof}
\end{proposition}

\setcounter{proposition}{8}
\begin{proposition}
    Let $\ms$ be a multiset of rule bases with $\ms=(\rb_1, ..., \rb_n)=(\{\F_1\cup\R\},...,\{\F_n\cup\R\})$ and let $r\in \R$. The problem of determining $|\{M \in \MI(\rb_i) \mid i\in \{1,\ldots,n\} r \in M\}|$ is $\#\textsf{P}$-complete.
    \begin{proof}
        Membership follows from Proposition~\ref{prop:mi} as deciding for a given $M$ whether $M\in \{M \in \MI(\rb_i) \mid i\in \{1,\ldots,n\}, r \in M\}$ is in $\textsf{P}$.
        
        The proof of $\#\textsf{P}$-hardness is analogous to the proof of Proposition~5 in \cite{Corea:2020a}. Observe that in the reduction the notion of ``issue'' coincides with notion of a minimal inconsistent subset containing the rule $\pi\leftarrow \alpha_1,\ldots,\alpha_n,\delta_1,\ldots,\delta_m$ if we add facts $a,\neg a$ for each atom $a$ occurring the in input instance.
    \end{proof}
\end{proposition}

\end{document}